\documentclass[letterpaper]{article} 
\usepackage{aaai25}  
\usepackage{times}  
\usepackage{helvet}  
\usepackage{courier}  
\usepackage[hyphens]{url}  
\usepackage{graphicx} 
\usepackage{natbib}  
\usepackage{caption} 
\nocopyright

\usepackage{amsmath}
\usepackage{amssymb}
\usepackage{multirow}
\usepackage{booktabs}
\DeclareMathOperator*{\argmax}{arg\,max}
\usepackage{algorithm}
\usepackage{algorithmic}

\usepackage{newfloat}
\usepackage{listings}
\DeclareCaptionStyle{ruled}{labelfont=normalfont,labelsep=colon,strut=off} 
\floatstyle{ruled}
\newfloat{listing}{tb}{lst}{}
\floatname{listing}{Listing}
\title{Hologram Reasoning for Solving Algebra Problems with Geometry Diagrams}
\author {
    Litian Huang\textsuperscript{\rm 1},
    Xinguo Yu\textsuperscript{\rm 1},
    Feng Xiong\textsuperscript{\rm 1},
    Bin He\textsuperscript{\rm 1},
    Shengbing Tang\textsuperscript{\rm 1},
    Jiawen Fu\textsuperscript{\rm 1}
}
\affiliations {
    \textsuperscript{\rm 1}Faculty of Artiﬁcial Intelligence in Education, Central China Normal University\\
    litianhuang@mails.ccnu.edu.cn, 
    xgyu@ccnu.edu.cn, 
    mountain@mails.ccnu.edu.cn,
    hebin@ccnu.edu.cn,
    tangshengbing@ccnu.edu.cn,
    fjw@mails.ccnu.edu.cn
}

\begin{document}

\maketitle

\begin{abstract}

    Solving Algebra Problems with Geometry Diagrams (APGDs) is still a challenging problem because diagram processing is not studied as intensively as language processing. To work against this challenge, this paper proposes a hologram reasoning scheme and develops a high-performance method for solving APGDs by using this scheme. To reach this goal, it first defines a hologram, being a kind of graph, and proposes a hologram generator to convert a given APGD into a hologram, which represents the entire information of APGD and the relations for solving the problem can be acquired from it by a uniform way. Then HGR, a hologram reasoning method employs a pool of prepared graph models to derive algebraic equations, which is consistent with the geometric theorems. This method is able to be updated by adding new graph models into the pool. Lastly, it employs deep reinforcement learning to enhance the efficiency of model selection from the pool. The entire HGR not only ensures high solution accuracy with fewer reasoning steps but also significantly enhances the interpretability of the solution process by providing descriptions of all reasoning steps. Experimental results demonstrate the effectiveness of HGR in improving both accuracy and interpretability in solving APGDs. 

\end{abstract}

%
\begin{links}
    \link{Code}{https://github.com/FerretDoll/HGR}
\end{links}

\section{Introduction}

\begin{figure}[t]
    \centering
    \includegraphics[width=0.9\columnwidth]{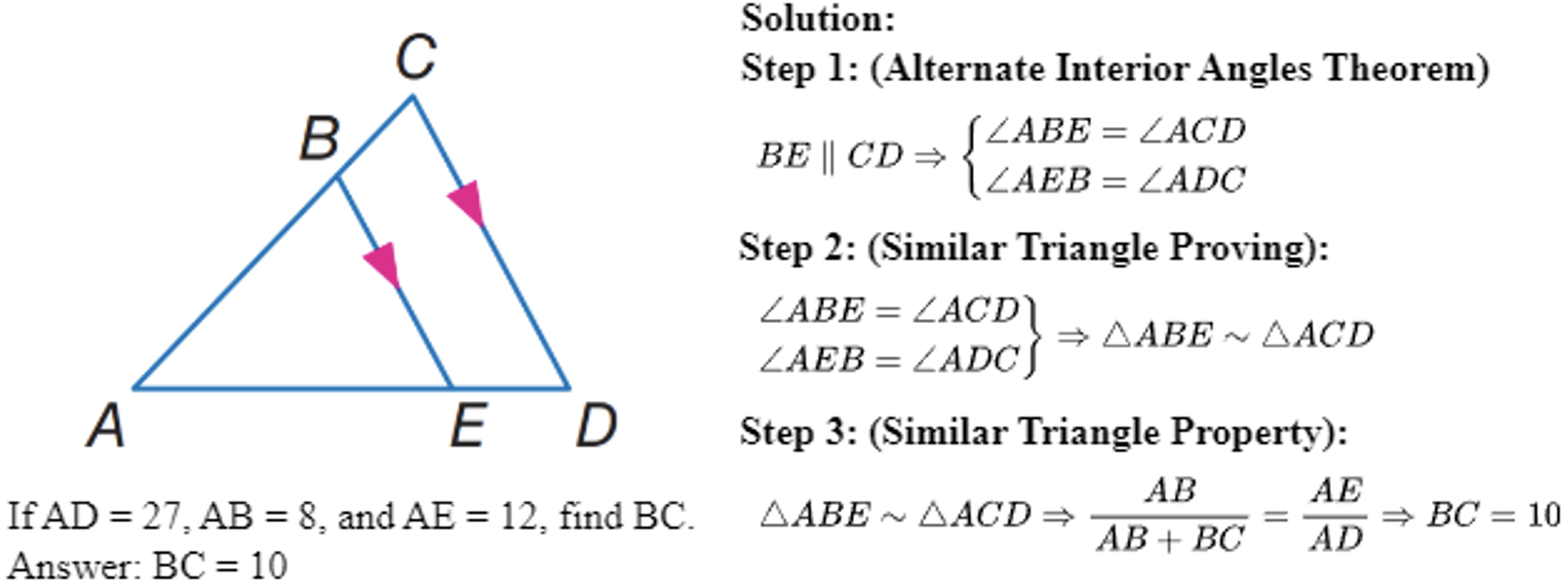}
    \caption{An example of an algebra problem with a geometry diagram, illustrating the use of geometric theorems to derive algebraic equations and find the final solution.}
    \label{fig1}
\end{figure}

Algebra Problems with Geometry Diagrams (APGDs) involve solving algebraic equations derived from the problem text and diagrams, which is a common task in educational contexts. These problems require solvers to interpret both textual descriptions and geometry diagrams, necessitating a combined understanding of algebraic and geometric principles \cite{Xia2021}. The complexity arises from needing to apply geometric theorems and manage implicit information within diagrams, which are not always explicitly stated (as shown in Fig. \ref{fig1}). Addressing these challenges is crucial for developing intelligent educational tools and advancing automated reasoning systems. Therefore, effective methods for APGD solving are essential for educational applications.

In the domain of solving APGDs, existing techniques are divided into two primary categories: the neural methods and the symbolic methods. Neural methods \cite{Chen2021,Chen2022,Ning2023,Liang2023} utilize neural networks to process multimodal inputs—integrating textual descriptions and geometry diagrams into a unified sequence for generating solutions. Recent advances in this area include the incorporation of Large Language Models (LLMs) like GPT-4V \cite{Lu2023mathvista,Kazemi2023}, which can handle complex language and diagrams, further enhancing the capability of neural models in understanding and solving APGDs. These methods are noted for their ability to capture the multimodal nature of the problems, providing a flexible approach to generating solutions. Meanwhile, symbolic methods focus on extracting and interpreting geometric relations from APGD, constructing symbolic representations that are essential for applying geometric theorems and algebraic operations. Symbolic methods include both traditional symbolic methods, which only rely on predefined rules and symbolic systems \cite{Seo2014,Seo2015,Huang2022,Huang2023}, and neuro-symbolic methods \cite{Lu2021,Peng2023,Wu2024}, which utilize neural models to enhance the accuracy and efficiency of symbolic systems. These methods emphasize the importance of accurately understanding the geometric context provided by diagrams, working in conjunction with textual description to solve APGDs.

Despite the progress made by existing methods, both neural methods and symbolic methods still have significant limitations. Neural methods \cite{Chen2021,Chen2022,Ning2023,Liang2023,Lu2023mathvista,Kazemi2023} often struggle with accurately interpreting mathematical diagrams and exhibit vulnerabilities in mathematical reasoning. This can result in errors when converting geometry diagrams into mathematical expressions. Additionally, the solutions often lack clarity in logical progression, making it difficult to follow the reasoning process. Traditional symbolic methods \cite{Seo2014,Seo2015,Huang2022,Huang2023}, while providing mathematical rigor, also have limitations. They are prone to generating redundant steps, resulting in longer solution times and decreased efficiency. To address these limitations, neuro-symbolic methods \cite{Lu2021,Peng2023,Wu2024} utilize neural networks to implement decision-making, integrating this with symbolic manipulation to enable step-by-step deduction. However, most of the existing neuro-symbolic methods are running on the same symbolic system as InterGPS \cite{Lu2021}, which necessitates rewriting code to add new theorems. This process can be inefficient and prone to errors, such as introducing bugs and creating inconsistencies, all of which can hinder the scalability and robustness of the system. Moreover, the existing methods typically only output the theorem names without specifying their application to particular geometric elements or the associated equations, which limits its utility in educational scenarios where detailed, interpretable outputs are crucial for effective teaching and learning.

To address these issues, we propose a novel method called HGR, a hologram-based reasoning method for solving APGDs. HGR parses the problem text and diagrams, converting them into a unified global hologram where vertices represent geometric primitives (points, lines, angles, etc.) and edges depict their relations. We maintain a model pool that contains multiple predefined graph models, each graph model uniquely characterized by a specific pattern hologram, effectively representing a theorem and corresponding reasoning rules. Deep reinforcement learning is employed to select the graph model from the pool. Once a graph model is selected and matched with the global hologram, it specifies which geometric theorems to apply to which primitives. The operations defined by the model are then applied to generate the corresponding algebraic equations and modify the global hologram by adding new vertices and edges or updating attributes. The process is repeated until a solution is reached, thereby providing a structured and efficient solution while enhancing interpretability.

Our contributions can be summarized as follows:

\begin{enumerate}
    \item We propose a hologram-based reasoning method that addresses APGD solving by unifying the information from problem text and diagrams into a hologram and performs reasoning directly on the hologram.
    \item We propose a method for effectively matching graph models within a hologram, supported by a specifically developed pool of models, to efficiently extract relevant geometric relations and algebraic equations.
    \item we propose a deep reinforcement learning method that optimizes the selection of graph models from the pool to enhance the efficiency of the model matching process.
\end{enumerate}

\section{Related Work}

\subsection{Methods of Solving APGDs}

Recent advancements in solving APGDs have primarily focused on two categories: neural methods and symbolic methods. Neural methods \cite{Chen2021,Chen2022,Ning2023,Liang2023} leverage neural networks to integrate textual and diagrammatic inputs, producing solutions through cross-modal representations. Neural methods excel at processing multimodal data, creating a unified framework for processing text and diagrams. However, they often struggle with accurately capturing the detailed aspects of diagrams and maintaining logical coherence in the solutions \cite{Trinh2024}. There is a growing trend in leveraging LLMs for solving APGDs, utilizing their extensive pre-trained knowledge to interpret and generate mathematical reasoning \cite{Lu2023mathvista,Kazemi2023}. However, these LLMs often struggle to accurately extract necessary information from geometry diagrams, leading to significant performance drops \cite{Zhang2024}. Their reliance on the quality and scope of training datasets also limits their generalization across different datasets and problem types, resulting in inconsistent performance and susceptibility to errors in mathematical reasoning \cite{Ahn2024}. Unlike neural methods, symbolic methods first parse the textual descriptions and geometric diagrams to extract structured representations, then utilize these representations to apply predefined theorems and rules for problem-solving. These methods \cite{Seo2014,Seo2015,Lu2021,Huang2022,Huang2023,Peng2023,Wu2024} demonstrate expertise in providing a clear rule-based framework for interpreting the geometric relations within APGDs, making them highly interpretable and ensuring mathematical rigor. However, symbolic methods are often constrained by their reliance on rigid symbolic systems based on formal language, limiting their applicability. To address these issues, our method utilizes a hologram-based reasoning system. It allows for more flexible and adaptable representations, improving the efficiency and accuracy of APGD-solving.

\subsection{Integration of Neural and Symbolic Methods}

To address the limitations of neural and traditional symbolic methods, the integration of symbolic and neural methods has led to significant advancements in solving APGDs. Inter-GPS \cite{Lu2021} and E-GPS \cite{Wu2024} use a trained theorem predictor that significantly improves the accuracy and efficiency of the symbolic solver. GeoDRL \cite{Peng2023} combines self-learning frameworks with symbolic manipulation, using Deep Q-Networks (DQN) \cite{Mnih2013} to choose theorems and guide deductive reasoning, thus merging interpretability with adaptability. AlphaGeometry \cite{Trinh2024}, a neuro-symbolic system that uses a neural language model to assist a symbolic deduction engine, effectively synthesizing and solving complex geometry-proving problems with human-readable proofs. These integrated methods leverage the strengths of both symbolic and neural methods, offering more comprehensive, efficient, and interpretable solutions for APGDs. In our method, we employ DQN to select graph models from the model pool, improving both the speed and accuracy of reasoning.

\begin{figure*}[htbp]
    \centering
    \includegraphics[width=0.9\textwidth]{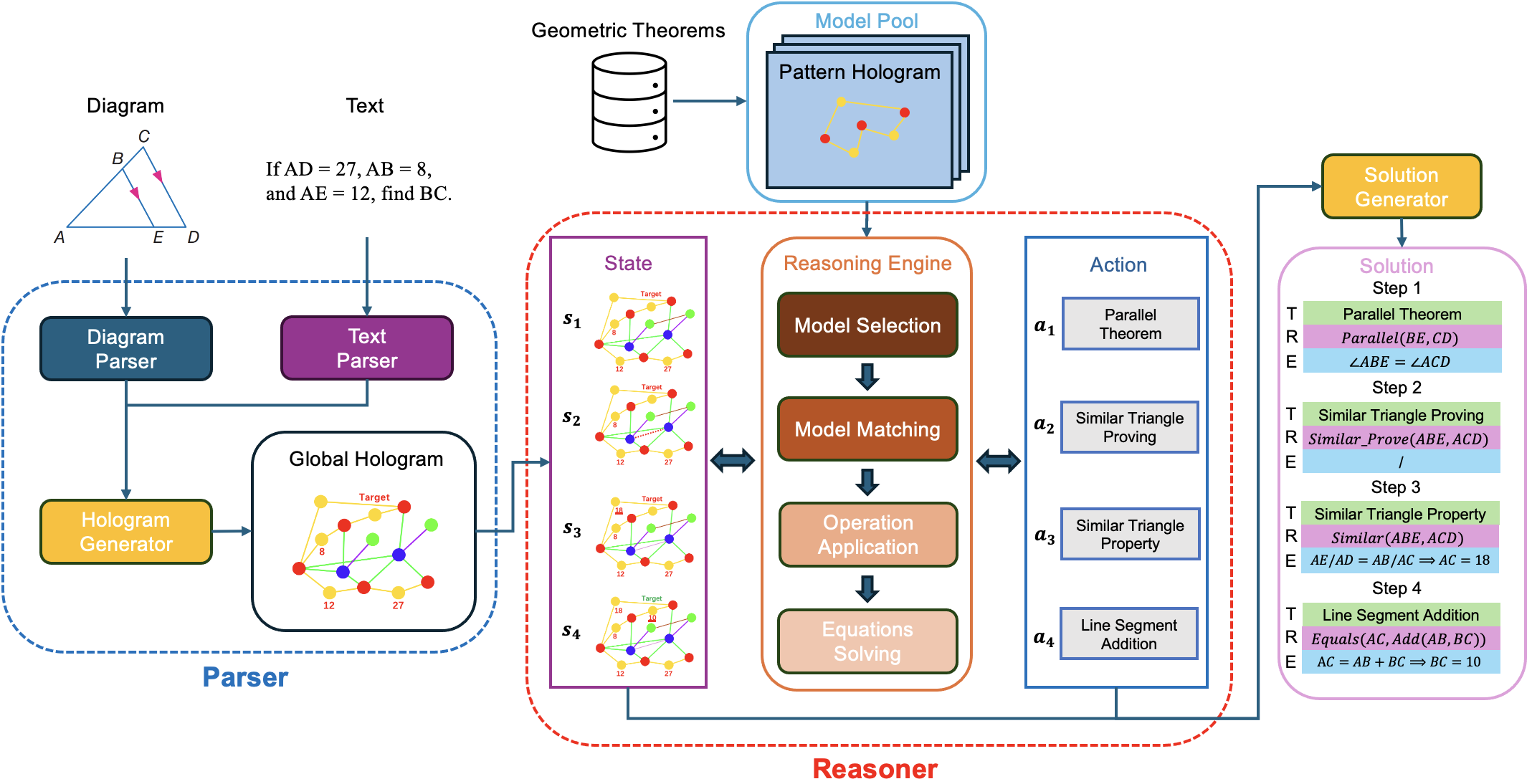}
    \caption{Overview of HGR which includes four components: 1) Parser, which processes the problem's text and diagram into a hologram; 2) Model pool, which contains a collection of predefined graph models representing different geometric theorems and reasoning patterns; 3) Reasoner, which selects appropriate graph models from the model pool, matches them to the global hologram, and applies operations to derive solutions; and 4) Solution generator, which outputs the final readable solution steps including Theorems (T), Relations (R), and Equations (E).}
    \label{fig2}
\end{figure*}

\section{Method}

The HGR method for solving APGDs consists of the parser, graph model construction and reasoner, as depicted in Fig. \ref{fig2}. In this section, the definition of the hologram and each HGR component are introduced separately.

\subsection{Hologram}

Inspired by the Geometry Logic Graph (GLG) used in GeoDRL \cite{Peng2023}, we optimize it to better suit the specific requirements of APGDs and propose the hologram which serves as the foundation for reasoning process and solution generation. 

The hologram is a heterogeneous attributed graph denoted as $G=(V,E,A,\hat{A})$, where:

\begin{itemize}
    \item \textbf{Vertices $V$}: The vertices in the hologram represent various geometric primitives, including points, lines, angles, arcs, circles and polygons.
    \item \textbf{Edges $E$}: The edges represent the geometric relations between the primitives, including adjacency, incidence (e.g., a point on a line), parallelism, perpendicularity, similarity and congruence. These edges ensure the hologram accurately reflects the spatial and relational data from the problem. The edges do not have a direction since the hologram is an undirected graph, reflecting the symmetrical nature of geometric relations.
    \item \textbf{Mathematical Attributes $A$}: Mathematical attributes associated with vertices include measurable properties such as lengths of lines, measures of angles, and area of polygons. Additionally, the hologram incorporates a special target attribute $\tau$ that specifies the problem's target. These attributes are critical for incorporating quantitative data necessary for problem-solving.
    \item \textbf{Visual Attributes $\hat{A}$}: Visual attributes associated with vertices are the properties of geometric primitives calculated from the diagram image, such as point positions, visual lengths and angles. Visual attributes are not used in the formal calculation process but serve as auxiliary data to support model matching process.
\end{itemize}

Using the hologram for reasoning in HGR provides several advantages. It provides a structured representation of the problem, effectively addressing the limitations of traditional symbolic reasoning methods, such as handling spatial and geometric relations inadequately. This structured approach enhances efficiency and precision, allowing for the rapid identification and application of relevant theorems and rules. Additionally, by avoiding the redundancy of symbolic grammar, it ensures that the reasoning process is clear and precise, facilitating better problem-solving in APGDs.

\subsection{Parser}

The parser of HGR plays a critical role in converting raw problem inputs into structured data that can be used for reasoning. The parser consists of three parts: text parser, diagram parser, and hologram generator.

\noindent \textbf{Text Parser}. The text parser adopts the rule-based approach used in Inter-GPS \cite{Lu2021} and GeoDRL \cite{Peng2023}. The text parser interprets the problem text $\mathcal{T}$ into a set of logical expressions in formal language (e.g., \textit{Triangle(A, B, C)} and \textit{PointLiesOnLine(A, Line(B, C))}). The parsing process involves identifying and extracting key elements from $\mathcal{T}$, including geometric primitives (e.g., points, lines, and angles), values, geometric relations and the problem's target. $\mathcal{T}$ is converted into a set of logical expressions, denoted as ${\Sigma}_{\mathcal{T}}$.

\noindent \textbf{Diagram Parser}. The diagram parser adopts an end-to-end model called PGDPNet \cite{Zhang2022} to analyze and interpret the geometry diagram $\mathcal{D}$ provided in the problem. PGDPNet takes the diagram image as input and efficiently identifies and extracts geometric primitives from $\mathcal{D}$. It also recognizes text annotations (e.g., lengths and angle measures) and geometric symbols (e.g., parallel and perpendicular indicators) within $\mathcal{D}$. These elements are then integrated into a set of logical expressions, denoted as ${\Sigma}_{\mathcal{D}}$.

\begin{figure}[t]
    \centering
    \includegraphics[width=0.9\columnwidth]{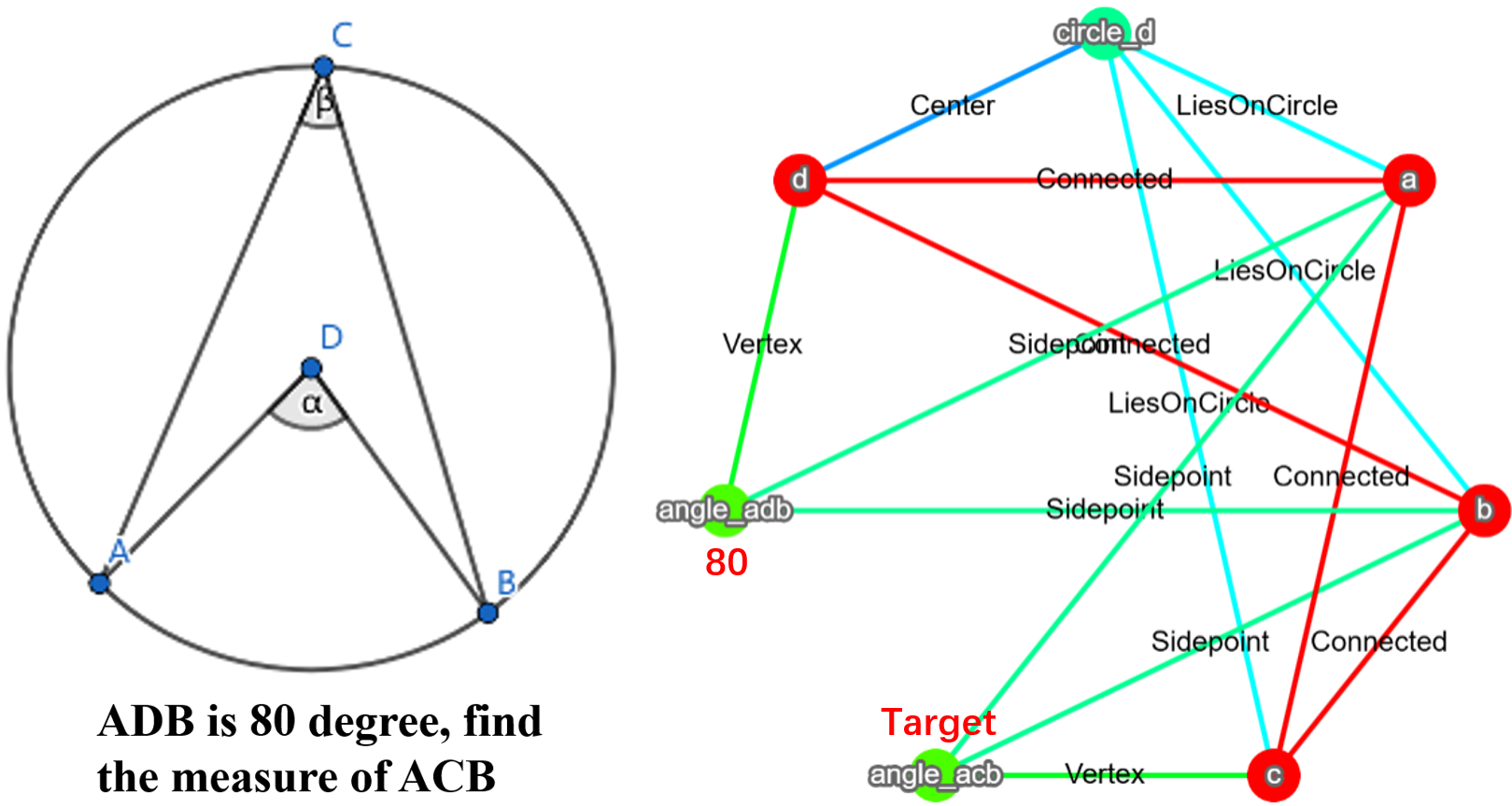}
    \caption{An example of an APGD with corresponding global hologram.}
    \label{fig3}
\end{figure}

\noindent \textbf{Hologram Generator}. The hologram generator is responsible for converting the parsed structured data $\Sigma = {\Sigma}_{\mathcal{T}}\cup{\Sigma}_{\mathcal{D}}$ into a hologram. Based on the hologram definition, the global hologram $G_{g}=(V_{g},E_{g},A,\hat{A})$ is generated, encompassing all geometric primitives and relations derived from the problem's given information. Fig. \ref{fig3} shows an example of an APGD and the global hologram.

\subsection{Graph Model Construction}

In HGR, graph models play a crucial role in the reasoning process, as they encapsulate geometric configurations and relations, enabling systematic application of theorems and rules to derive algebraic equations essential for solutions. These models are divided into two primary types: \textbf{Proving Models} $\mathcal{M}_{prov}$ and \textbf{Property Models} $\mathcal{M}_{prop}$. Both types of models share common components but serve different purposes within the reasoning process. The collection of these models forms the model pool, denoted as $\mathcal{M}={\mathcal{M}}_{prov}\cup{\mathcal{M}}_{prop}$.

\begin{figure}[t]
    \centering
    \includegraphics[width=0.9\columnwidth]{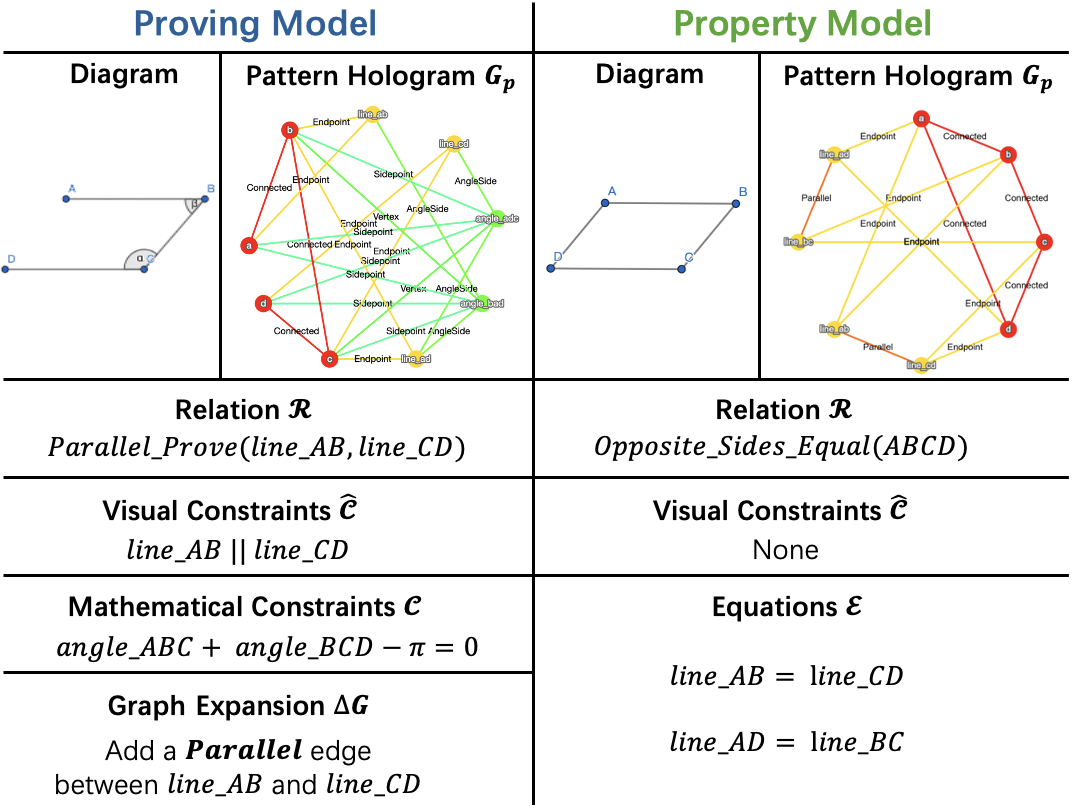}
    \caption{Examples of proving model and property model, showing the distinct roles and applications of each type.}
    \label{fig4}
\end{figure}

Proving model ${m}_{prov}=(G_{p},\mathcal{R},\hat{\mathcal{C}},\mathcal{C},\Delta G)\in \mathcal{M}_{prov}$ and property model $m_{prop}=(G_{p},\mathcal{R},\hat{\mathcal{C}},\mathcal{E})\in \mathcal{M}_{prop}$ have the following common components: 1) \textbf{Pattern Hologram $G_{p}=(V_{p},E_{p})$} is generated based on the hologram definition without including attribute values for the vertices. It is used in both ${m}_{prov}$ and $m_{prop}$ to match with $G_{g}$. It identifies specific geometric configurations corresponding to known theorems or properties. 2) \textbf{Relation $\mathcal{R}$} represents the geometric relation template associated with each graph model. While they do not actively participate in the reasoning process, they play a crucial role in the solution-generation process by providing a readable explanation of the solution. 3) \textbf{Visual Constraints $\hat{\mathcal{C}}$} are crucial for ensuring the accuracy of pattern matching in the hologram, as the matching process itself primarily ensures topological consistency. $\hat{\mathcal{C}}$ provides auxiliary information necessary for resolving ambiguities that arise from purely topological matches. For instance, when two triangles are known to be similar but the corresponding angles are not explicitly identified, $\hat{\mathcal{C}}$ helps determine the correct angle correspondences by comparing visual aspects like the proximity of angle measures. This ensures that the correct geometric relations are established, thus enhancing the precision of the reasoning process.

Proving model ${m}_{prov}$ contains the following specific components: 1) \textbf{Mathematical Constraints $\mathcal{C}$} are used to verify whether the mathematical attributes of geometric primitives in $G_{g}$, such as angle measures or line segment lengths, meet the conditions required for the theorem being established. 2) \textbf{Graph Expansion $\Delta G$} refers to a set of operations applied to $G_{g}$, which involves adding new vertices and edges and modifying mathematical attributes. It facilitates the application of geometric theorems and extends the reasoning process by introducing new geometric primitives or modifying existing ones.

Property model ${m}_{prop}$ contains the specific component \textbf{Equations $\mathcal{E}$}, which describes the quantitative properties of geometric primitives. The generated equations are then added to the equation set for solving.

Fig. \ref{fig4} illustrates how the proving and property models are used in HGR. The proving model ${m}_{prov}$ demonstrates the process of proving that two lines are parallel by verifying that the sum of co-interior angles equals 180 degrees. It then updates $G_{g}$ by adding a \textit{Parallel} edge. This updated hologram allows the property model ${m}_{prop}$ to match and establish geometric relations, such as identifying that opposite sides of a parallelogram are equal. The detailed matching and reasoning processes are elaborated in \textbf{Reasoner}.

\subsection{Reasoner}

\begin{figure}[t]
    \centering
    \includegraphics[width=0.9\columnwidth]{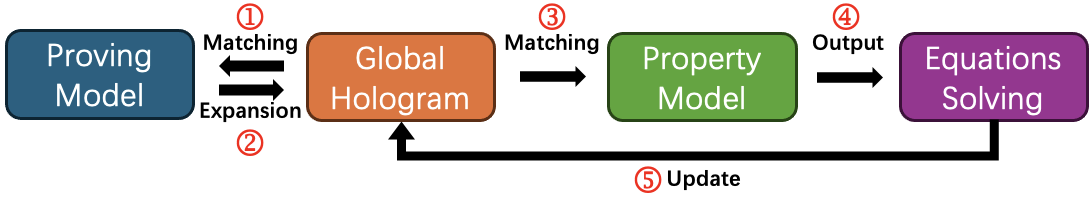}
    \caption{The iterative reasoning process using the proving model and property model.}
    \label{fig5}
\end{figure}

The reasoner in HGR iteratively applies geometric theorems to the global hologram $G_{g}$ until the final solution is reached. As illustrated in Fig. \ref{fig5}, the reasoning process involves several key steps: 1) \textbf{Proving model matching}, where the proving model is matched with $G_{g}$; 2) \textbf{Graph expansion}, updating $G_{g}$ with new vertices or edges based on the matching; 3) \textbf{Property model matching}, where the property model is matched with $G_{g}$; 4) \textbf{Equations derivation}, where equations are derived based on the matched property models; and 5) \textbf{Attributes update}, updating the mathematical attributes of the vertices of $G_{g}$ based on the solutions obtained from the equations. The iterative process continues until the target answer is obtained, ensuring all relevant theorems are applied accurately.

\noindent \textbf{Model Matching} (Step 1 and 3). The model matching process of the reasoner determines which theorems to apply to specific geometric primitives identified in $G_{g}$. The model matching process consists of graph pattern matching, mathematical constraints verification (for proving models) and visual constraints verification.

Graph pattern matching involves finding a mapping function between the pattern hologram $G_{p}=(V_{p},E_{p})$ and global hologram $G_{g}=(V_{g},E_{g},A,\hat{A})$:

\begin{equation}
    \label{eq1}
    M:V_{p} \to V_{g}
\end{equation} 

\noindent where $V_{p}$ and $V_{g}$ represent the sets of vertices in $G_{p}$ and $G_{g}$ respectively. The function $M$ must preserve the graph's structure by maintaining the presence and absence of edges between corresponding vertice pairs.

For this task, we utilize the VF3 algorithm \cite{Carletti2017}, which has been shown to offer high efficiency and accuracy in graph pattern matching \cite{Carletti2020}.

Following the graph pattern matching, the mathematical constraints verification ensures that the mathematical attributes $A$ of the mapped vertices meet the specific conditions required for the theorems to be applied. Each mathematical constraint $c_i \in \mathcal{C}$ is evaluated as follows:

\begin{align}
    \label{eq2}
    c_{i}(M({V}_{p})) &= 
    \begin{cases}
        \text{True} & f_{i}(A(M({V}_{p})))=0 \\
        \text{False} & \text{otherwise}
    \end{cases}
\end{align}

\noindent where $M({V}_{p})$ denotes retrieving the mapped vertices from $V_{g}$ based on the mapping function $M$, $A(\cdot)$ denotes obtaining the mathematical attributes of vertices from $A$, and $f_{i}(\cdot)$ denotes evaluating the $i$-th algebraic function of $\mathcal{C}$.

The overall mathematical constraints verification $\mathcal{C}$ is determined by the logical conjunction of all constraints $c_{i}$:

\begin{equation}
    \label{eq3}
    \mathcal{C}(M(V_{p})) = \bigwedge_{i=1}^{n} c_i(M(V_{p}))
\end{equation}

If any constraint $c_{i}$ fails, $\mathcal{C}$ outputs \textbf{False}, indicating that the theorem cannot be applied to the current state. The visual constraints verification follows a similar process, with $c_i \in \mathcal{C}$ and $A$ replaced by $\hat{c}_i \in \hat{\mathcal{C}}$ and $\hat{A}$ respectively. When graph pattern matching, mathematical constraints verification (for proving models), and visual constraints verification all succeed, the graph model is considered successfully matched. The geometric relation template is instantiated using $M$ for solution presentation, denoted as $\mathcal{R}(M(V_{p}))$.

\noindent \textbf{Operations Application} (Step 2 and 4). Once the model matching process is successful, specific operations are executed depending on the type of graph model.

For the proving model, graph expansion is performed based on the mapping $M$ established during the model matching, resulting in a new global hologram $G^{'}_{g}$:

\begin{equation}
    \label{eq4}
    G^{'}_{g} \gets G_{g} + \Delta G(M({V}_{p}))
\end{equation}

\noindent where $\Delta G(M({V}_{p}))$ denotes the set of graph expansion operations applied to $G_{g}$.

For the property model, variables of equations $\mathcal{E}$ are replaced with the actual mathematical attributes of $V_{g}$ from $A$:

\begin{equation}
    \label{eq5}
    \tilde{\mathcal{E}} = \{ e_i \in \mathcal{E} \mid e_i = g_i(A(M(V_{p}))) \}
\end{equation}

\noindent where $g_i(\cdot)$ denotes replace variables of the $i$-th equation in $\mathcal{E}$. The generated set of equations $\tilde{\mathcal{E}}$ is then added into the equation set $\mathcal{E}_{total} = \mathcal{E}_{total} \cup \tilde{\mathcal{E}}$ for the equations solving process.

\noindent \textbf{Equations Solving and Attributes Update} (Step 5). The set of equations $\mathcal{E}_{total}$ is solved to find the unknown values. The solution is then used to update the mathematical attributes in $G_{g}$. If the updated mathematical attributes satisfy the problem target $\tau$, the reasoning process concludes. Otherwise, the reasoning process continues iteratively.

\noindent \textbf{Enhanced Model Selection}. In HGR, a critical challenge is efficiently selecting the appropriate graph models to match. A naive method would involve employing a heuristic strategy to exhaustively search through all graph models, but this method becomes inefficient as the complexity and number of graph models increase.

To address this, we design a model selection agent that implements deep reinforcement learning to select the graph model at each reasoning step. The objective is to find a deterministic \textit{policy} $\pi :s\rightarrow \pi (s) =a$ that maximize the expected cumulative rewards:

\begin{equation}
    \label{eq6}
    \pi^*=\argmax_{\pi}\sum_{t=0}^{\infty}\mathbb{E}\left[{r(s_t,a_t)}\right]
  \end{equation}

\noindent where $r(s_t,a_t)$ is the reward of obtained at step $t$ for taking action $a_t$ in state $s_t$, and $\pi^*$ is the optimal policy.

In the implementation, the \textit{state} $s$ is represented by the global hologram $G_{g}$ and the \textit{action} $a$ is represented by selecting the graph model $m \in \mathcal{M}$. The GraphTransformer \cite{Yun2019} is utilized to encode the state $s$, capturing its structural and attribute information, and to assess the effectiveness of potential actions.

The reward function $r(s,a)$ for selecting a model $m$ based on the global hologram $G_{g}$ is defined as follows:

\begin{align}
    \label{eq7}
    r(s,a) &= 
    \begin{cases}
        1.0-{\alpha e}^{- \frac{\theta}{\sigma }} & s^{'} \vdash \tau\\
        -{\alpha e}^{- \frac{\theta}{\sigma }} & s^{'} \neq s \text{ and } s^{'} \nvdash \tau\\
        -1.0 & s^{'}=s
    \end{cases}
\end{align}

\noindent where $s^{'}$ is the updated global hologram after taking action $a$, $\theta$ is the time spent, ${\alpha e}^{- \frac{\theta}{\sigma }}$ is a time penalty factor. The agent gets the positive reward if $s^{'}$ satisfies the target $\tau$.

The optimization process is carried out using the DQN algorithm \cite{Mnih2013}, which guides the agent in efficiently navigating through the model pool and selecting the most promising graph model for matching at each step. Algorithm \ref{alg:reasoning} details the iterative reasoning process for solving the APGD using either a heuristic strategy or a model selection agent.

\begin{algorithm}[tb]
    \caption{Iterative Reasoning Process with Different Model Selection Strategy}
    \label{alg:reasoning}
    \textbf{Input}: Global hologram $G_g$, model selection strategy (Heuristic or Agent)\\
    \textbf{Parameter}: Graph models pool $\mathcal{M}={\mathcal{M}}_{prov}\cup{\mathcal{M}}_{prop}$\\
    \textbf{Output}: Problem target $\tau$
    \begin{algorithmic}[1] 
    \STATE Initialize $\mathcal{E}_{total} \gets \emptyset$
    \WHILE{$\tau$ not satisfied}
        \IF{Heuristic strategy}
            \FOR{each $m_{prov} \in \mathcal{M}_{prov}$}
                \IF{$m_{prov}$ matches $G_g$}
                    \STATE $G_g \gets G_g + \Delta G$
                    \STATE \textbf{break}
                \ENDIF
            \ENDFOR
            \FOR{each $m_{prop} \in \mathcal{M}_{prop}$}
                \IF{$m_{prop}$ matches $G_g$}
                    \STATE Derive $\tilde{\mathcal{E}}$
                    \STATE $\mathcal{E}_{total} \gets \mathcal{E}_{total} \cup \tilde{\mathcal{E}}$
                    \STATE \textbf{break}
                \ENDIF
            \ENDFOR
        \ELSE
            \STATE $s \gets \text{GraphTransformer}(G_g)$
            \STATE $m \gets \pi(s)$
            \IF{$m$ is proving model}
                \IF{$m$ matches $G_g$}
                    \STATE $G_g \gets G_g + \Delta G$
                \ENDIF
            \ELSIF{$m$ is property model}
                \IF{$m$ matches $G_g$}
                    \STATE Derive $\tilde{\mathcal{E}}$
                    \STATE $\mathcal{E}_{total} \gets \mathcal{E}_{total} \cup \tilde{\mathcal{E}}$
                \ENDIF
            \ENDIF
            \ENDIF
        \STATE Solve $\mathcal{E}_{total}$
        \STATE Update attributes of $G_g$
    \ENDWHILE
    \STATE \textbf{return} $\tau$
    \end{algorithmic}
    \hrule 
\end{algorithm}

\begin{table*}[t]
    \centering
    \begin{tabular}{l|c|cccc|ccccc}
    
    \toprule[1.5pt]
    \multirow{2}{*}{\textbf{Method}}           & \multicolumn{1}{c|}{\multirow{2}{*}{All}} & \multicolumn{4}{c|}{Problem Target}                                                        & \multicolumn{5}{c}{Problem Type}                                                                                 \\ \cline{3-11} 
                                               & \multicolumn{1}{c|}{}                     & Angle                & Length               & Area                 & Ratio                 & Line                 & Triangle             & Quad                 & Circle               & Other                \\ \hline
    Human                                      & 56.9                                      & 53.7                 & 59.3                 & 57.7                 & 42.9                  & 46.7                 & 53.8                 & 68.7                 & 61.7                 & 58.3                 \\
    Human Expert                               & 90.9                                      & 89.9                 & 92.0                 & 93.9                 & 66.7                  & 95.9                 & 92.2                 & 90.5                 & 89.9                 & 92.3                 \\ \hline
    FiLM \cite{Perez2018}                                      & 31.7                                      & 28.7                 & 32.7                 & 39.6                 & 33.3                  & 33.3                 & 29.2                 & 33.6                 & 30.8                 & 29.6                 \\
    FiLM-BERT \cite{Devlin2019}                                  & 32.8                                      & 32.9                 & 33.3                 & 30.2                 & 25.0                  & 32.1                 & 32.3                 & 32.2                 & 34.3                 & 33.3                 \\
    FiLM-BART \cite{Lewis2020}                                 & 33.0                                      & 32.1                 & 33.0                 & 35.8                 & 50.0                  & 34.6                 & 32.6                 & 37.1                 & 30.1                 & 37.0                 \\ \hline
    Inter-GPS \cite{Lu2021}                                 & 57.5                                      & 59.1                 & 61.7                 & 30.2                 & 50.0                  & 59.3                 & 66.0                 & 52.4                 & 45.5                 & 48.1                 \\
    Inter-GPS (GT)                             & 78.3                                      & 83.1                 & 77.9                 & 62.3                 & 75.0                  & 86.4                 & 83.3                 & 77.6                 & 61.5                 & 70.4                 \\ \hline
    GeoDRL \cite{Peng2023}                                    & 68.4                                      & 75.5                 & 70.5                 & 22.6                 & 83.3                  & 77.8                 & 76.0                 & 62.9                 & 53.8                 & 48.1                 \\
    GeoDRL (GT)                                & 89.4                                      & 86.5                 & \textbf{93.7}                 & 75.5                 & 100.0                 & 87.7                 & \textbf{93.1}                 & 90.2                 & 78.3                 & 77.8                 \\ \hline
    HGR (ours)                             &           68.7                                &      78.0                &         65.2             &           38.9           &           87.5            &          75.7            &            72.5          &          64.8            &           53.5           &         71.8             \\
    \textbf{HGR (GT)}                      & \textbf{89.6}                                      & \textbf{89.7}                 & 89.9                 & \textbf{87.2}                 & \textbf{100.0}                 & \textbf{92.4}                 & 90.9                 & \textbf{90.2}                & \textbf{81.9}                 & \textbf{83.5}                  \\ \bottomrule[1.5pt]
    \end{tabular}
    \caption{Accuracy (\%) results of HGR and compared baselines on Geometry3K. GT refers to the use of ground truth parsing results, rather than those generated by the parser.}
    \label{tab1}
\end{table*}

\setlength{\tabcolsep}{2mm}
\begin{table}[]
    \centering
    \begin{tabular}{l|c|c}
    \toprule[1.5pt]
    \textbf{Method} & Overall Acc. (\%)      & Avg. Step (T/R/E)                             \\ \hline
    InterGPS (GT)                & 78.3          & 6.68/-/-                            \\ 
    GeoDRL (GT)                  & 89.4          & 2.37/-/-                      \\ 
    HGR (GT)                 & \textbf{89.6} & \textbf{2.26/7.29/17.55}        \\ \bottomrule[1.5pt]
    \end{tabular}
    \caption{Evaluation results of HGR and compared baselines on Geometry3K. Avg. Step (T/R/E) means the average number of theorems/relations/equations to solve each problem. Avg. Time means the average time spent on each solved problem.}
    \label{tab2}
\end{table}

\section{Experiments}

\subsection{Datasets and Evaluation Metrics}

\noindent \textbf{Datasets}. The experiments are primarily conducted on Geometry3K \cite{Lu2021}, a comprehensive dataset designed for evaluating geometry problem-solving systems. Geometry3K consists of 3,002 SAT-style APGDs. Each problem in the dataset is presented as a single-choice question with four choices, accompanied by problem text, a geometric diagram, and explicit parsing annotations in a formal language. The problems cover a wide range of geometric shapes, including lines, triangles, circles, quadrilaterals, and other polygons, providing a robust benchmark for evaluating the performance of problem solvers.

\noindent \textbf{Evaluation Metrics}. To evaluate the performance in solving APGDs, two primary metrics are used: 1) \textbf{accuracy}: the accuracy is determined by whether the numerical result produced is closest to the correct answer among four choices. In cases where the method fails to produce a numerical result, a random selection is made;  2) \textbf{reasoning steps}: the average number of geometric theorems applied to generate the solution. For HGR, theorems are counted by the number of successful graph model matches. Additionally, the count includes the number of geometric relations and equations generated, represented as a combined metric: \textit{theorems} / \textit{relations} / \textit{equations}.

\subsection{Baselines}

HGR is compared against several baseline methods recognized for their effectiveness in solving APGDs. FiLM \cite{Perez2018} is used for visual reasoning on abstract images, adapted for APGD solving. Enhanced versions, FiLM-BERT \cite{Devlin2019} and FiLM-BART \cite{Lewis2020}, incorporate BERT and BART encoders to improve performance. Both Inter-GPS \cite{Lu2021} and GeoDRL \cite{Peng2023} utilize the same symbolic system for solving APGDs, with Inter-GPS employing deep learning and GeoDRL leveraging deep reinforcement learning to enhance reasoning accuracy and efficiency.

\subsection{Implementation Details}

For the model construction, a graph model pool containing 13 proving models and 51 property models representing different geometric theorems is prepared. To enable efficient graph pattern matching using the VF3 algorithm, both the global hologram and pattern holograms are converted into GRF format. For training the model selection agent, both the heuristic strategy and random selection are employed to generate model selection samples for the training set, with the heuristic strategy focused on generating more samples with positive rewards, while the random selection tends to generate more samples with negative rewards. The agent is then pre-trained on these samples for 5000 steps to establish an initial policy. Finally, the agent is trained using DQN on the training set to refine its policy and improve its performance.

\subsection{Evaluation}

The evaluation of HGR's accuracy on the Geometry3K dataset, as detailed in Table \ref{tab1}, demonstrates its superior capabilities in solving APGDs compared to baseline methods. Specifically, HGR achieves an overall accuracy of 68.7\%, which improves to 89.6\% when using Ground Truth (GT) parsing results. This performance is comparable to the state-of-the-art GeoDRL model and surpasses Inter-GPS, particularly in solving problems related to area calculations or circle type.

In terms of efficiency, HGR outperforms the baselines by requiring fewer reasoning steps on average. As shown in Table \ref{tab2}, HGR requires only 2.26 average steps compared to 2.37 for GeoDRL, highlighting its effectiveness in generating concise and accurate solutions.

\subsection{Ablation Study}
\setlength{\tabcolsep}{2mm}
\begin{table}[]
    \centering
    \begin{tabular}{l|c|c}
    \toprule[1.5pt]
    \multirow{2}{*}{\textbf{Method}} & Overall       & Avg. Step   \\
                                     & Acc. (\%)     &  (T/R/E)  \\ \hline
    HGR (GT)                     & \textbf{89.6} & 2.26/7.29/17.55  \\ \hline
    -w/o agent                       & 83.4          & 2.84/8.65/19.79 \\
    -w/o proving model               & 79.5          & 2.35/7.81/19.03  \\
    -w/o property model              & 27.0          & 0.03/0.08/0.00 \\
    -w/o $\mathcal{C}$                           & 73.6          & 1.41/5.82/13.24 \\
    -w/o $\hat{\mathcal{C}}$                             & 63.1          & 1.06/3.98/8.70  \\ \bottomrule[1.5pt]
    \end{tabular}
    \caption{Results of ablation study on Geometry3K. -w/o agent means replacing the model selection agent with the heuristic strategy. -w/o proving model and -w/o property model mean removing proving models and property models from the model pool respectively. -w/o $\mathcal{C}$ and -w/o $\hat{\mathcal{C}}$ mean removing mathematical constraints and visual constraints in the model matching respectively.}
    \label{tab3}
\end{table}

In the ablation study, we evaluated the impact of various components in HGR on Geometry3K. As presented in Table \ref{tab3}, the results demonstrate the significance of each component. The removal of the model selection agent results in a notable decrease in overall accuracy from 89.6\% to 83.4\%, while also increasing the average reasoning steps significantly. This highlights the agent's crucial role in efficiently selecting the appropriate graph models. The removal of proving models reduces accuracy to 79.5\%, highlighting their crucial role in modifying the global hologram to enable successful matching of property models. The removal of property models drastically reduces accuracy to 27.0\%, highlighting their critical role in generating the necessary algebraic equations for solving problems. Additionally, the exclusion of mathematical constraints and visual constraints lead to declines in accuracy to 73.6\% and 63.1\%, respectively. These results demonstrate that both constraints are critical for ensuring the correctness and validity of the model matching process.

\subsection{Discussion}

\noindent \textbf{Interpretability of HGR}. Table \ref{tab4} compares the interpretability of various APGD solvers, showing that HGR offers the most comprehensive interpretability. UniGeo \cite{Chen2022}, as a neural method, is only capable of generating the final computational sequence, lacking detailed intermediate explanations. Inter-GPS and GeoDRL, while utilizing the same symbolic system, can output the sequence of theorems used and the geometric primitives to which these theorems were applied. However, their generated equations only include the numerical values of geometric primitives, making it difficult to interpret the underlying geometric relations. Besides, they are unable to generate human-readable solutions directly. In contrast, HGR not only demonstrates the specific geometric theorems applied to corresponding geometric primitives at each step, but also generates equations that demonstrate the algebraic relations between primitives. Additionally, through the use of graph model relation templates, HGR provides human-readable solutions, making it highly suitable for educational applications. 
\setlength{\tabcolsep}{1.5mm}
\begin{table}[t]
    \centering
    \begin{tabular}{c|cccc}
    \toprule[1.5pt]
    \multirow{2}{*}{\textbf{Method}} & \multicolumn{4}{c}{Interpretability}                                                                     \\ \cline{2-5} 
                            & \multicolumn{1}{c|}{Theorem} & \multicolumn{1}{c|}{Primitive} & \multicolumn{1}{c|}{Equation} & Solution \\ \hline
    UniGeo                  & \multicolumn{1}{c|}{}        & \multicolumn{1}{c|}{}          & \multicolumn{1}{c|}{}         & \checkmark      \\
    Intet-GPS               & \multicolumn{1}{c|}{\checkmark}     & \multicolumn{1}{c|}{\checkmark}       & \multicolumn{1}{c|}{}         &          \\
    GeoDRL                  & \multicolumn{1}{c|}{\checkmark}     & \multicolumn{1}{c|}{\checkmark}       & \multicolumn{1}{c|}{}         &          \\
    HGR                 & \multicolumn{1}{c|}{\checkmark}     & \multicolumn{1}{c|}{\checkmark}       & \multicolumn{1}{c|}{\checkmark}      & \checkmark      \\ \bottomrule[1.5pt]
    \end{tabular}
    \caption{Comparison of interpretability of APGD solvers.}
    \label{tab4}
\end{table}

\begin{figure}[t]
    \centering
    \includegraphics[width=0.9\columnwidth]{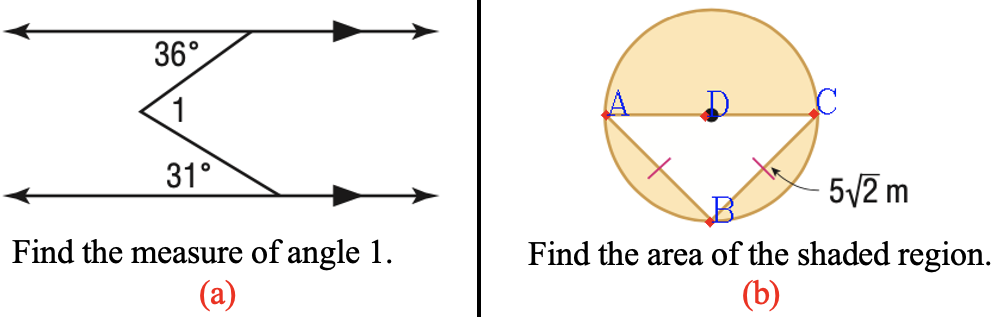}
    \caption{Failure examples of HGR.}
    \label{fig6}
\end{figure}

\noindent \textbf{Failure Case}. HGR currently lacks the ability to solve two types of problems, which contributes to a reduction in overall accuracy. The first type involves problems that require auxiliary constructions, such as adding auxiliary points, lines, or circles, as shown in Fig. \ref{fig6}(a). The second type of failure case is related to problems that involve calculating the area of shaded regions, illustrated in Fig. \ref{fig6}(b). These limitations highlight specific aspects for future improvement, particularly in enhancing the model's ability to handle auxiliary constructions and more complex area calculations.

\section{Conclusion and Future Work}

This paper has presented HGR, a high-performance method for solving APGDs in which the hologram reasoning scheme played the main contribution in improving solving accuracy and computing efficiency and enhancing the interpretability of the solution. During this process, this paper has three technique contributions. First, it proposed the hologram and the method of converting a given APGD into a hologram, which possesses the excellent property that a procedure can acquire all the relations from it for solving the problem. Second, it proposed a model-matching method to acquire the relations from the hologram and it prepared a pool of graph models for acquiring relations. Third, it proposed a deep reinforcement learning method to select models from the pool to speed up the model matching process. 

Three future works can be done on the base of this paper. First, the method presented in this paper can be improved in hologram construction and relation acquisition from hologram. Second, we want to transfer the hologram reasoning into other types of problems. Third, we want to integrate the method for solving APGDs and the method for solving word algebra problems into a uniform method.

\bibliography{aaai25}

\end{document}